\documentclass[conference]{IEEEtran}

\IEEEoverridecommandlockouts                              

\usepackage{algorithm}
\usepackage[noend]{algpseudocode}

\usepackage[utf8]{inputenc}
\usepackage{graphicx,color}
\usepackage{cite}
\usepackage{caption}
\usepackage{subcaption}
\usepackage{times}  
\usepackage{epstopdf}
\usepackage[hyphens]{url}  
\usepackage{mathtools, nccmath}

\usepackage{multirow}

\usepackage{array}
\usepackage{hyperref}
\usepackage[bottom]{footmisc}
\usepackage{tabu}
\usepackage{booktabs}
\usepackage{amsmath,amsfonts,amssymb}

\title{\Large \bf
Scaffolding Reflection in Reinforcement Learning Framework\\ for Confinement Escape Problem
}

\author{Nishant Mohanty and Suresh Sundaram
\thanks{Nishant Mohanty and Suresh Sundaram are with the Department
of Aerospace Engineering, Indian Institute of Science, Bangalore,  India. (e-mail: \tt\small (nishantm,vssuresh)$@$iisc.ac.in).}%
}

\begin{document}

\maketitle
\thispagestyle{empty}
\pagestyle{empty}

\begin{abstract}
In this paper, a novel Scaffolding Reflection in Reinforcement Learning (SR2L) is proposed for solving the confinement escape problem (CEP). In CEP, an evader's objective is to attempt escaping a confinement region patrolled by multiple pursuers. Meanwhile, the pursuers aim to reach and capture the evader. The inverse solution for pursuers to try and capture has been extensively studied in the literature. However, the problem of evaders escaping from the region is still an open issue. The SR2L employs an actor-critic framework to enable the evader to escape the confinement region. A time-varying state representation and reward function have been developed for proper convergence.  The formulation uses the sensor information about the observable environment and prior knowledge of the confinement boundary. The conventional Independent Actor-Critic (IAC) method fails to converge due to sparseness in the reward. The effect becomes evident when operating in such a dynamic environment with a large area. In SR2L, along with the developed reward function, we use the scaffolding reflection method to improve the convergence significantly while increasing its efficiency. In SR2L, a motion planner is used as a scaffold for the actor-critic network to observe, compare and learn the action-reward pair. It enables the evader to achieve the required objective while using lesser resources and time. Convergence studies show that SR2L learns faster and converges to higher rewards as compared to IAC. Extensive Monte-Carlo simulations show that a SR2L consistently outperforms conventional IAC and the motion planner itself as the baselines.

\end{abstract}

\section{Introduction}

The recent developments in robotics and sensing technologies have created significant interest among researchers to deploy them for various territory surveillance problems. In particular, three problems have gained attention: territory protection, pursuit-evasion, and confinement escape. Territory protection requires multiple robots to tackle intruders cooperatively \cite{Analikwu2017MultiagentLI,8988241}. In pursuit-evasion, an agent tries to capture the evader before it escapes a given region \cite{1067989,7251450}. A closely related problem called the reach-and-avoid game is a class of planar multi-agent pursuit-evasion games. Unlike pursuit-evasion games, a team of evaders aims to avoid capture by a team of pursuers while simultaneously attempting to reach a target location \cite{8279644,8823952,ZHOU201828}.
For confinement escape problems (CEP) \cite{LI2016442}, the evaders attempt to escape from confinement without being captured by the pursuers. The formulation of CEP is an essential proxy for many robotic applications. In this problem, robots with varying degrees of cooperation and observation attempt to achieve designated objectives. It results in a high dimensional problem, thus increasing the complexity of obtaining an optimal CEP solution.

In literature, the first formulation for a typical confinement escape problem was given by \cite{LI2016442}. It defines a circular confinement region and uses artificial forces between the agents to control the pursuers and the evaders. \cite{LeeWi1}, \cite{7503136} provide further analysis of escape time with different initialization and the required winning conditions. These papers restrict the pursuers' motion to move along the circular boundary, thus providing a sub-optimal solution.  Related work like \cite{8718039} analytically formulates a control strategy for an evader to reach a given target location. It uses a subset of the evader's state for a nonlinear state feedback control while removing any restrictions on the pursuers. However, the proposed region-based control strategy performs well for a small number of pursuers only. The aforementioned works assume the evader has information regarding the entire region, thus causing dimensionality issues as the number of pursuers increases. Also, analytical approaches use handcrafted features to get a solution. As all the features cannot be captured in a formulation, it may lead to a sub-optimal solution.  One need to develop a new approach to overcome above-mentioned issues.



In this paper, a novel scaffolding reflection in reinforcement learning (SR2L) is proposed. SR2L employs an actor-critic framework for an evader to escape the confinement against multiple dynamic pursuers. A time-variant reward function is proposed for the evader to learn and extract essential features from the proposed time-variant state representation for an optimal solution. The evader uses information about the observable vicinity and their prior knowledge about the static boundary. This helps in avoiding the dimensionality issues and can work for any number of pursuers. Here, pursuers are free to move in any direction within the region. The evader learns to maneuver and escape from the confinement by using the proposed reward function.
However, for the evader to learn in an uncertain and complex environment of CEP would require a significant amount of time. The limited success conditions present in CEP lead to sparse reward scenarios. In SR2L, a well-defined reward function and a motion planner (MP) are used to reduce the effect of spareness. In literature, MP with learning algorithms often leads to overall efficiency being dependent on the motion planner itself. Here, the SR2L framework is designed such that the overall efficiency is independent of it.
The role of MP is to decrease the randomization during searching in a large environment. The network exploits human knowledge (in the form of MP) and tends to exceed its efficiency while also achieving the level much faster. In the literature, the potential field method (PFM) has been extensively used as a motion planner in confinement escape problems (CEP) \cite{LeeWi1,7503136}. In this paper, the PFM is used for scaffolding the evader's learning process. The scaffold helps the evader reflect upon its action by observing the probable actions suggested by the PFM algorithm. Extensive Monte-Carlo simulations have been performed to show the efficiency of a trained SR2L evader against the baseline, i.e., independent actor-critic  (IAC) and the PFM methods individually.
The convergence study of SR2L against IAC shows that the former converges $15$ times faster as compared to the latter. Also, SR2L converges to a higher average reward (approximately $-0.15$) where as IAC converges to $-0.25$.

The remaining paper's organization is as follows: Section \ref{sec:related_works} presents the related works concerning the SR2L framework. Section \ref{sec:prob_def} presents the mathematical formulation of the confinement escape problem. Section
\ref{sec:sr2l} describes the decision-making framework and the proposed learning process (SR2L). Section \ref{sec:eval} discusses the performance evaluation using Monte Carlo simulations and comparison with the conventional methods. Finally, the conclusions from the studies are summarised in section \ref{sec:conclusion}.

\section{Related Works}
\label{sec:related_works}
In recent times, the literature has addressed problems that involve decision-making and control by modeling them as Markov's games. Games are used as test-beds to accurately capture many of the ideas, planning, and reasoning. In literature, deep reinforcement learning (DRL) has been widely used for building agents to learn complex control for competitive games. For example, a human-level agent for playing Atari games is trained with deep Q-networks \cite{atari}. The use of supervised learning and self-play for training enabled the machine to beat a human professional in the game of Go \cite{Silver16}. Further, a more generalized DRL method is applied to the game of Chess, and Shogi \cite{Silver2017MasteringCA}. It led to an increasing interest in using DRL-based algorithms for tasks involving sequential decision making. Since then, it has been implemented to solve the sequential problems in the fields of social sciences, and engineering \cite{DBLPjournals}. For example, in \cite{hu2018reinforcement}, the trajectory design problem is formulated and solved using the proposed multiple robots decentralized Q-learning method.

DRL methods seem promising; however, with the increasing complexity of the environment of operation, these methods become difficult to train, and the time taken for the same is higher. Also, it requires careful consideration for formulating a reward function and designing the neural network architecture. Training a network can fail even for simple tasks if rewards are sparse in nature, i.e., if the success condition of the objective is a rare occurrence \cite{Bellemare2016UnifyingCE}. It is generally the case for navigation in large spaces. In \cite{inproceedings123} this problem addressed by introducing a proxy reward function that is less sparse than the actual objective. However, poorly derived proxy rewards can lead to agents learning to exploit the reward function and terminate the episode early rather than the actual objective. Formulating a good reward function is well understood for some areas like 1v1 games \cite{atari}, but for tasks like navigation, it remains challenging and adds to training difficulties. Thus, in this work, a similar ideology is used. A proxy reward function is developed to emulate the fundamental objective by embedding the sensor information available to the agent.

To reduce the training time for DRL and to increase its performance efficiency, \cite{yamada2020motion} used a motion planner (MP) and a model-free reinforcement learning (actor-critic) for implementation of a task. A similar method is used in the paper \cite{Chiang2018LearningNB} for training in the presence of static obstacles. However, these works consider using the MP to expand the agent's available action space, thus creating a joint action space. In this case, if a sampled action from the RL policy is in the original action space, an agent directly executes the action to the environment; otherwise, the motion planner computes a path to move the agent to faraway points for execution. 
Even though it increases overall efficiency, it is heavily dependent on the motion planner's performance. In this paper, SR2L is developed such that the action space of MP and actor-critic are not augmented; hence the overall efficiency is independent of MP performance. The purpose of MP is to act as a scaffold for the actor-critic network to reflect upon its actions, compare and learn.


\section{Confinement Escape Problem Definition}
\label{sec:prob_def}

\begin{figure}
\centering
\includegraphics[width=0.9\columnwidth]{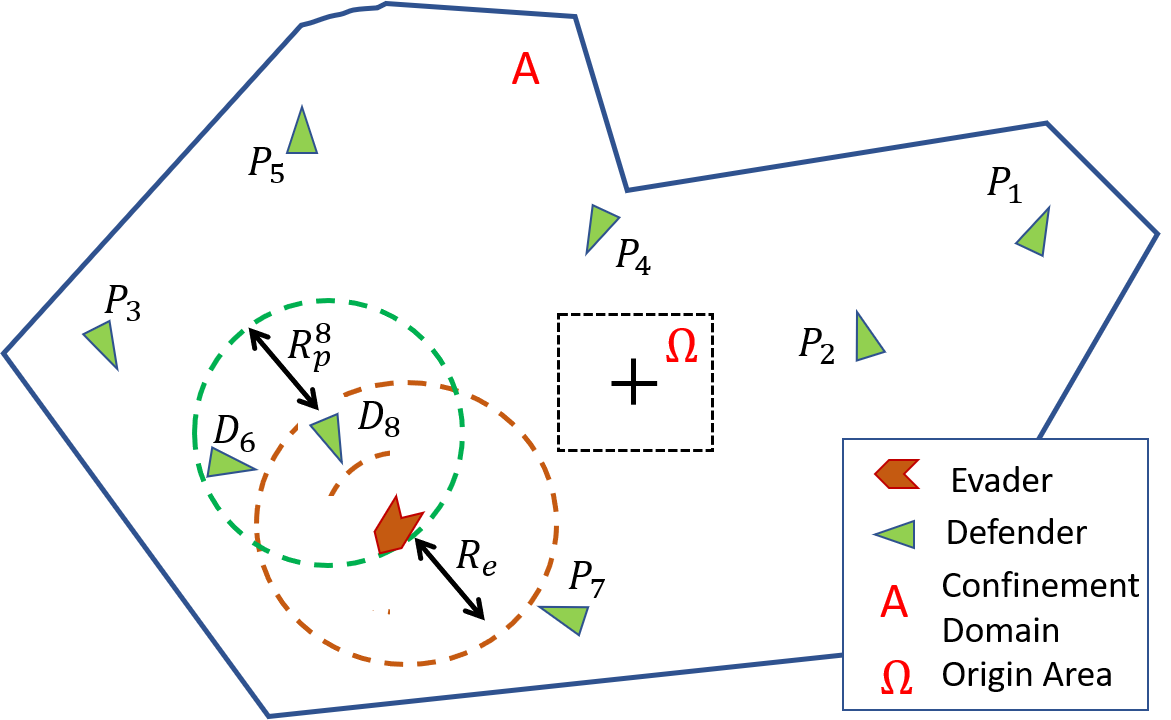}
\caption{Environment for CEP, with evader and multiple pursuers.}
\label{arena}
\end{figure}

In this section, the confinement escape problem (CEP) is formulated for the evader ($E$), with $m$ pursuers ($P_1, P_2, \cdots, P_m$) patrolling the closed confinement region $\bf{A} \subset \Re^2$. A typical confined region with pursuers and evader is shown in fig. \ref{arena}. $\Omega$ is the origin region where the evader $E$ is initialized, and the pursuers are distributed randomly in the entire region $\bf{A}$ that is outside of $\Omega$. The boundary of $\bf{A}$ is closed and can be any shape. The objective of the evader $E$ is to escape from the closed confinement region without being captured by the pursuers. The pursuers search for the evader without any communication between them and try to capture or restrict the evader from escaping. The game is terminated if the evader escapes the confinement within a given time period $t^{max}$, or is captured by a pursuer. For example, in fig. \ref{arena}, $8$ pursuers are initialized within the region $A$ outside of $\Omega$ and are chasing the evader $E$. The position of the $i^{th}$ pursuer is represented as $X^i_p=[x^i_p,y^i_p]$. The pursuers are initialized with different velocities, i.e., the velocity of $i^{th}$ pursuer is $V^i_p$. The velocities are restricted such that $V_p^{min} \leq V^i_p \leq V_p^{max}$. The heading direction of the $i^{th}$ pursuer $\psi_p^i$ is randomly initialized, and the pursuer is free to change the heading direction without any constraint.  Using the above definitions, the motion of the $i^{th}$ pursuer can be guided by the following kinematic equations,
\begin{eqnarray}
    \dot{x}_{p}^i &=& V_{p}^i cos(\psi_{p}^i)\\
    \dot{y}_{p}^i &=& V_{p}^i sin(\psi_{p}^i)
\end{eqnarray}
where, $\dot{x}_{p}^i$ and $\dot{y}_{p}^i$ represents the velocity of $i^{th}$ pursuer along cartesian $x$ and $y$ axes. 

The pursuers are heterogeneous in nature, i.e., the velocity and sensing radius is different for every pursuer. Let us assume that every pursuer is equipped with a LIDAR sensor to detect the evader, whose range is $R^{i}_p$. The pursuers maintain their velocity and heading until it detects an evader or boundary of the confinement region. Upon reaching the boundary, the pursuers get deflected away from the boundary, thus changing its heading while maintaining the velocity. If a pursuer detects the evader, it starts chasing by changing its heading towards the evader while moving with maximum velocity $V^{max}_{e}$. In this game, $\delta r$ distance is considered to be the capture radius, i.e., if the distance between the evader and any pursuer is less than $\delta r$, it is considered as a successful capture.    

For the evader $E$ to escape the confinement region, it uses a LIDAR of range $R_e$  to sense its nearby environment and a GPS sensor for localization. It is assumed that the evader has information about the boundary and can find the distance to every point on the boundary from the current position. The position of $E$ with respect to the origin is $X_e=[x_e,y_e]$. Let the velocity of the evader be $V_e$, and the heading be $\psi_e$. The maximum velocity of the evader is considered to be higher than that of the pursuers, i.e., $V' = V^{max}_e/V^{max}_p > 1$. The evader motion is guided by the following equations,
\begin{eqnarray}
    \dot{x}_{e} &=& V_{e} cos(\psi_{e}) \\
    \dot{y}_{e} &=& V_{e} sin(\psi_{e})
\end{eqnarray}
where, $\dot{x}_{e}$ and $\dot{y}_{e}$ represents the velocity of the evader along Cartesian $x$ and $y$ axes. In practice, the evader is initialized in the origin region $\Omega$ with an initial velocity of $0$ m/s, along with a random heading angle $\psi_{e}$. Also, no pursuer is initialized in the region $\Omega$. 

It is assumed that, upon sensing, the evader can detect the pursuers' velocity and heading direction. Also, the evader is equipped with a localization sensor to detect their position accurately. The evader aims to achieve a motion such that it avoids the pursuers and reaches the nearest boundary point. Assuming $R_b$ to be the maximum distance of the boundary from the point of initialization, the objective function for $E$ in its simplified form, at any given time, can be given as,
\begin{equation}
    L = min \left\{  \sum_{i=0}^{m} \frac{R_e - d_i}{m*R_e} 
    +  \frac{d_b}{R_b} \right\}
    \label{eq:obj}
\end{equation}
where $d_i$ is the distance from the $i^{th}$ pursuer among $m$ detected by the evader and  $d_b$ is the distance of the evader from the nearest point of the boundary. Hence, the mission is for the evader to escape the confinement region by avoiding the pursuers while minimizing the time taken in the process.

\section{Scaffolding Reflection in Reinforcement Learning for CEP}
\label{sec:sr2l}
In this section, first, the actor-critic formulation and the reinforcement learning framework of CEP are given. Following that, the learning procedure followed by the description of the scaffolding reflection in reinforcement learning (SR2L) is presented.

\subsection{Actor-critic formulation for CEP}
\label{sec:actor_critic_formulation}
In a reinforcement learning framework, the CEP is formulated as a markov decision process. Here, the process is described by an experience tuple represented as $<s, a, r, s'>$. At any given time, $s$ represents the state of the evader in the environment. The evader interacts with the environment by choosing an action $a \subset \Re$ using a policy $\pi(a|s)$. The evader's interaction with the environment leads to the next state $s'$ while gaining a reward $r$. The evader learns an optimal policy, $\pi^*(a|s)$, which gives the utility of the action space for its observed state. The goal of the evader is to learn a policy that maximizes the received discounted rewards. Here, $\gamma \in (0, 1]$ is the discount factor that determines how much the policy favors immediate reward over long-term rewards.

In this paper, an actor-critic deep reinforcement learning framework is used because of its ability to learn continuous action space.
Conventionally, for the actor, a policy gradient method is used to update actions so that actions with higher expected rewards have higher utility value for an observed state. Policy gradient techniques \cite{sutton} estimates the gradient of expected utility values with given policy parameters $\theta$ using,
\begin{equation}
    \nabla_\theta  J (\pi_\theta)  =\nabla_\theta \log \left(\pi_\theta (a_t|s_t)\right)G(t)
    \label{eq:grad}
\end{equation}
where, $G(t)$ is the cumulative reward. Actor-critic methods \cite{konda} increase the stability by replacing cumulative rewards with a function approximation of the expected rewards. In this paper, the function approximation formulation (Q-value function) estimates expected discounted returns in a given state-action pair as,
\begin{equation}
    Q_\psi(s_t,a_t) = \mathbb{E}\left[ \sum_{t'=t}^{\infty} r_{t'} (s_{t'},a_{t'})\right]
\end{equation}
Here, soft actor-critic \cite{Haarnoja2018SoftAO} have been used to focus not only on maximizing long term rewards but also maximizing the entropy of the policy. The term entropy here refers to the the measure of unpredictability of a random variable. Maximum entropy RL learns a soft value function by modifying the policy gradient to incorporate an entropy term. Thus redefining $G(t)$ as:
\begin{equation}
    G(t) = Q_\psi(s,a) - b(s) -a \log(\pi_\theta(a|s))
    \label{eq:modifed_gradient}
\end{equation}
where $b(s)$ is called the baseline for the Q-value function and is a function of state $s$. Actor critic learning is then achieved by minimizing the regression loss function given as:
\begin{equation}
    \mathbb{L}_Q(\psi) = \mathbb{E}_{(s,a,r,s')}
    \left[(Q_\psi(s',a,) - y)\right]
    \label{eq:loss}
\end{equation}
\begin{equation}
    y = r(s,a) + \gamma \mathbb{E} [Q_{\psi'}(s',a') -a \log(\pi_\theta(a|s))]
    \label{eq:modified_target}
\end{equation}
Here, $Q_{\psi'}$ refers to the target Q-value function and $y$ is the target value.
As the motion of the evader is not the same in every episode, the utilities can vary drastically with episodes. This causes $G(t)$ to give high variance in rewards. Updating policy with these rewards leads to high variance in the logarithm of the policy distribution. Thus resulting in noisy gradients (eqn. \ref{eq:grad}). It causes instability while learning, along with slow and non-optimal convergence of policy gradient methods.

\subsection{State and reward definition for CEP}




For the reinforcement learning framework to achieve the desired results, it is required to develop a reward function to emulate the objective function (eqn. \ref{eq:obj}) of the evader. Hence, the point cloud data from the LIDAR is used by the evader to avoid capture by pursuer. To generate the point cloud, the LIDAR is assumed to have a resolution of $N_s$. Every point in the point cloud is associated with two values, i.e., the range measurements ($z_i \leq R_e$) and the angle from the evader frame of reference. For the point cloud data to be encoded into the reward function, it is represented as a list,
\begin{equation}
    D_l = \{ K_s*(z_i/R_e) \}_{i=1}^{N_s}
\end{equation}
where, $K_s$ is a proportionality constant.
This data helps the evader to escape the pursuers and prevents itself from being captured.
Similarly, for the evader to get out of the confinement, the evader is assumed to have information about the boundary of the confinement. Thus, with the help of a localization sensor on-board, the evader can calculate the distance ($b_i$) from the boundary points in all directions. To maintain the size of the data similar to that of LIDAR, the resolution considered here is also $N_s$. This data is represented as a list,
\begin{equation}
    D_b = \{ K_s*(1 - (b_i/R_b)) \}_{i=1}^{N_s}
\end{equation}
where, $K_s$ is the proportionality constant. The same constant is used for both $D_l$ and $D_b$, to represent the data in the same scale. 
The evader requires to escape the confinement as quickly as possible, before the termination time $t^{max}$ of the game. Therefore, the evader requires the information regarding the time elapsed in a given game as an input to the system. The time elapsed information is embeded in to the state of the evader using a time factor $t_f$  and is defined as,
\begin{equation}
\label{eq:timefactor}
    t_f = (1 - (t/t^{max}))/2
\end{equation}
where, $t$ is the amount of time elapsed in a given game. The final state of the evader, at any given time, is taken to be a weighted sum of the sensor information with the time factor as a proportionality constant. It is given as,
\begin{equation}
    s = t_f*(W_l* D_l + W_b * D_b)/(W_l + W_b) 
    \label{eq:state}
\end{equation}
where, $W_l$ and $W_b$ represents the weights for LIDAR sensor and boundary information respectively.

In a reinforcement framework, agent-environment interaction provides a reward $r$ for the agent to train. To emulate the objective function (eqn. \ref{eq:obj}) while also ensuring escape in minimum time, the reward function used in this paper embeds information about distance from the boundary, the location of observed opponents, and time elapsed to capture all essential features of the environment. It is done by designing rewards for each component, followed by a weighted sum for the overall reward.
First, the rewards due to the evaders interaction with the pursuers can be given using the following set of equations,
\begin{eqnarray}
    W_i &=& (1 - (d_i^{t}/R_e))\\
    r_d &=& \sum_{i=0}^{m-1} W_i*(V_{rel}^{max}*dt - (d_i^{t} - d_i^{t-1}))
\end{eqnarray}
where, $d_i$ is the distance from the $i^{th}$ pursuer among $m$ detected by the evader in the vicinity. Also $V_{rel}^{max} = V_e^{max} - V_p^i\cos(\theta_i)$ and $dt$ is the time step.
Here, $V_p^i$ is the the velocity of the $i^{th}$ detected pursuer and $d_i^{t}$ distance from it at time $t$. And, $V_p^i\cos(\theta_i)$ is the pursuers velocity towards the evader, as $\theta_i$ is the angle between the heading of the pursuer and the line joining it to the evader. The formulation for $r_d$ evaluates the evader's performance compared to the maximum possible distance it can travel away from the pursuer in a time step $dt$. The weight $W_i$ ensures higher weightage for the pursuers closer to the evader.
Similarly, with the information about the boundary, the evader's interaction with the boundary leads to a reward given by,
\begin{equation}
    r_b = V_{e}^{max}*dt - (d_b^{t-1} - d_b^{t})
\end{equation}
where, $d_b^t$ is the distance of the evader from the nearest boundary point at a given time $t$. The formulation of $r_b$ evaluates the evader's action by comparing it with the maximum possible distance the evader can travel towards the boundary in a time step $dt$.
Finally, to generate the reward function used in this work, the time factor, formulated in eqn. \ref{eq:timefactor}, is included as a proportional constant to provide lesser positive reward as the game proceeds with time. Thus, forcing the evader to learn to escape the confinement in minimum time. The final reward formulation for the evader's interaction with the environment can be given as,
\begin{equation}
    r = t_f*[\{(1 - \sum_{i=0}^{m-1} W_i)/(1 + m)\}*r_b + r_d]
    \label{eq:reward}
\end{equation}
As mentioned in section \ref{sec:actor_critic_formulation}, in every iteration upon generation of the state $s$ (using eqn. \ref{eq:state}) is generated. The state $s$ acts as input to the actor-critic network and the output of the network is the $Q$ values. The network selects an action $a$ corresponding to the maximum $Q$ value. This action enables the agent to interact with the environment thus generating the next state $s'$ (using eqn. \ref{eq:state}) and the reward $r$ (using eqn. \ref{eq:reward}). Thus, an experience tuple is created $<s, a, r, s'>$. The actor-critic networks are then trained by minimizing the regression loss given by eqn. \ref{eq:loss} and \ref{eq:modified_target}. In this method, as no additional algorithm has been used except the actor-critic framework, from here onwards, it is mentioned to be as independent actor-critic (IAC). Even after using shaped reward functions that emulate the true reward, the convergence characteristics remain slow \cite{vargas2020effects} due to the sparseness. To improve the convergence even in the presence of sparseness, multi-agent setup has been used by \cite{christianos2021shared}. It uses experience gathered by all the agents to train a single policy. However, such a setup will not work in this case due to the presence of a single evader. Hence, there is a need to develop a framework to ensure faster convergence of the learning algorithm in sparse reward conditions.

\subsection{Scaffolding Reflection in Reinforcement Learning}

\begin{figure}
\centering
\includegraphics[width=0.9\columnwidth]{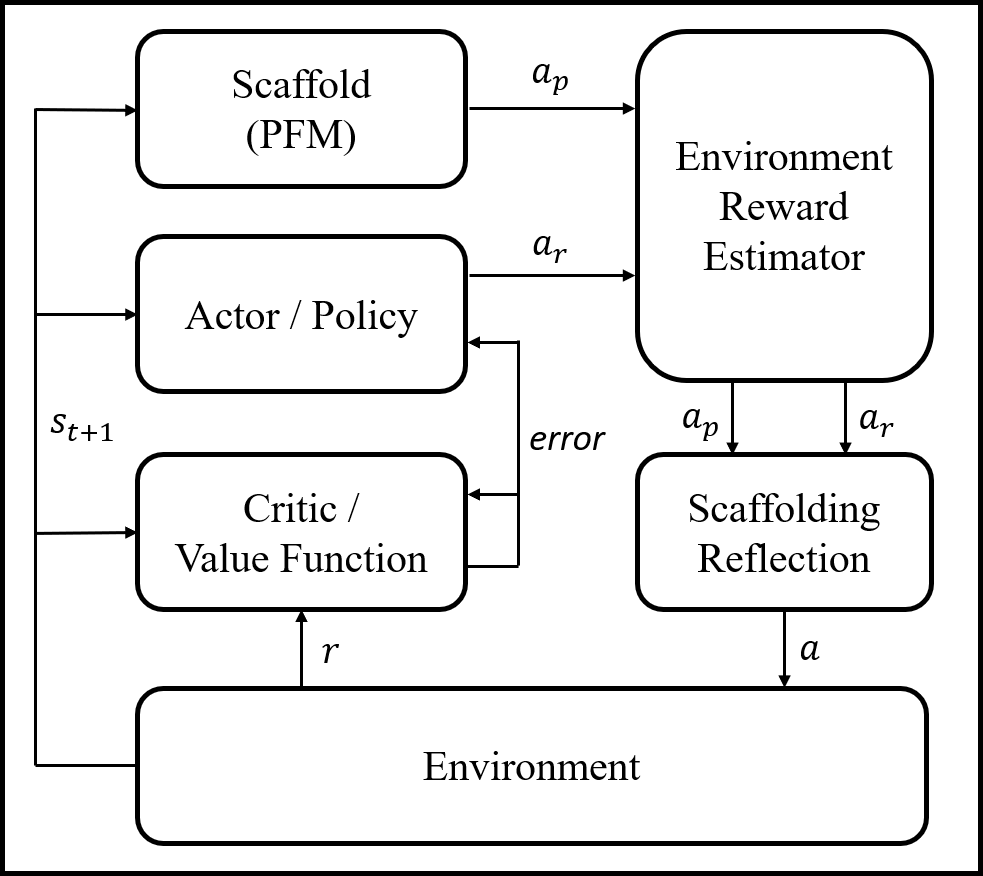}
\caption{The schematic diagram of scaffolding reflection in actor-critic framework for CEP.}.
\label{fig:schematic}
\end{figure}

\begin{algorithm}

\caption{Pseudocode of SR2L Algorithm}
\begin{algorithmic}

\Statex
 \State $s$ \,\, $\leftarrow$ state of the evader using eqn. \ref{eq:state};
 \State $a_p$ $\leftarrow$ output from PMF for the given state $S$;
 \State $a_r$ $\leftarrow$ output from Actor network for given state $S$;
 \State $s'_p$\, $\leftarrow$ estimated next state due to $a_p$;
 \State $s'_r$\, $\leftarrow$ estimated next state due to $a_r$;
 
 \State $r_p$ $\leftarrow$ expected reward for $s \rightarrow s'_p$ transition;
 \State $r_r$ $\leftarrow$ expected reward for $s \rightarrow s'_r$ transition;

 \State $D_f$ $\leftarrow$ percentage difference between expected rewards is calculated using eqn \ref{eq:diff};
 
 \State $\beta$ $\leftarrow$ a positive threshold value is assigned
 \If{$D_f > -\beta$}:
    \State $a$ $\leftarrow$ $a_r$;
    \State where, $a$ is final action taken by evader;
    
    \State $r$ $\leftarrow$ $r_r$;
    \State where, $r$ is final reward to the evader;
    
\Else:    
    \State $a$ $\leftarrow$ $a_p$;
    \State where, $a$ is final action taken by evader;
    
    \State $r$ $\leftarrow$ $r_r - \Delta r$;
    \State where, $r$ is final reward to the evader;
    \State and, $\Delta r \leftarrow |r_p - r_r|$;     
 \EndIf
 \State $s'$\, $\leftarrow$ next state of evader due to $A$.
 \State $<s,a,r,s'>$\, $\leftarrow$ final experience tuple.
 \State Train actor-critic network using $<s,a,r,s'>$.
 
\end{algorithmic}
\label{Algo}
\end{algorithm}

In this section, we present a novel Scaffolding Reflection in Reinforcement Learning (SR2L) algorithm to overcome the sparsity in the reward due to dynamic and large environments. This method uses an actor-critic framework along with a motion planner (MP). The MP here acts as a scaffold for the actor-critic to reflect upon its action and learn accordingly. Fig. \ref{fig:schematic} shows the schematic diagram of the proposed SR2L framework. First, the state representation is generated by encoding the information about the observable vicinity and the boundary information. It then acts as input to the actor-critic (AC). The action suggested by it and that of the motion planner is compared in the environment reward estimator, and the action-reward pair is generated. It enables the actor-critic to observe, compare and learn with the help of a motion planner.  Earlier works use the potential field method (PFM), which uses artificial forces between the agents to control the evaders in CEP \cite{LI2016442}. In this paper, it is used as the motion planner to act as a scaffold for actor-critic as it is computationally less intensive.
For PFM, an artificial repulsive force from the pursuers and attractive force towards the boundary's closest point is applied. The net force can be given as,
\begin{equation}
    F_{net} = \{\sum_{i=0}^{m} (k_p/d_i^2). \hat{d_i} \} + (k_b/d_b^2).\hat{d_b}   
\end{equation}
where, $\hat{d_i}$ is the unit vector away from the opponents and $\hat{d_b}$ is the unit vector towards the nearest boundary point. $k_p$ and $k_b$ are proportional constants for pursuers part and boundary part respectively. 

\begin{figure*}
\centering
    \subcaptionbox{time step = 25.\label{fig:a}}
    {\includegraphics[width=0.329\textwidth]{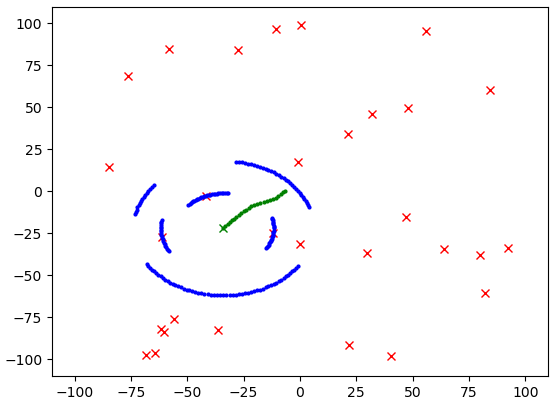}}
    \subcaptionbox{time step = 50\label{fig:b}}
    {\includegraphics[width=0.329\textwidth]{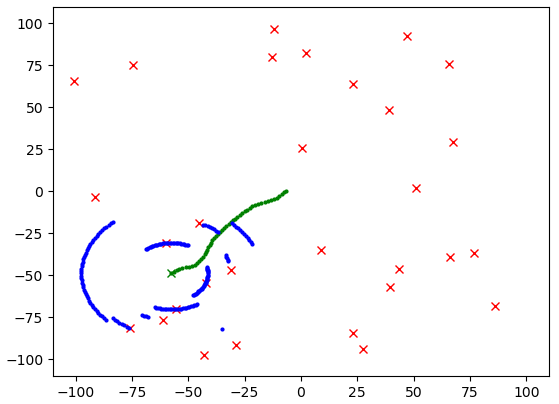}}
    \subcaptionbox{time step = 80.\label{fig:c}}
    {\includegraphics[width=0.329\textwidth]{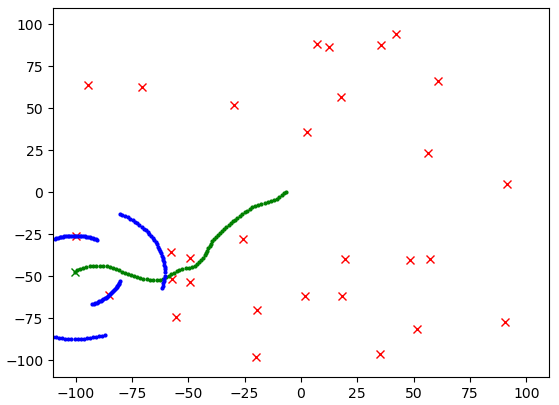}}
    
    \caption{A sequence of actions taken by the evader to escape the rectangular confinement region of dimensions $200m\times200m$. The green trajectory represents the path followed by the evader, whereas the blue arcs represents the point cloud data.}
    \label{fig:action_sequence}
\end{figure*}

The formal description of the SR2L methodology is given in Algorithm \ref{Algo}. The state action pair is decided by considering the action decision by both PFM ($a_p$) andactor-critic($a_r$) network along with their corresponding rewards ($r_p$ and $r_r$). During training, in every time step, the rewards are compared to quantify the performance of the actor-critic network against PFM. It is formulated as a percentage of the difference between the rewards and is given by,
\begin{equation}
    D_f = \{(r_r - r_p)/|r_p + \epsilon|\}*100 \%
    \label{eq:diff}
\end{equation}
A scaffolding condition is established, which decides the final reward to be awarded to the actor. It is formulated as,
\newcommand{\twopartdef}[4]
{
	\left\{
		\begin{array}{ll}
			#1 & \mbox{if } #2 \\
			#3 & \mbox{if } #4
		\end{array}
	\right.
}
\begin{equation}
    \label{eq:scaf_reward}
    r = \twopartdef { r_r } {D_f \geq -\beta} {r_r - |(r_p - r_r)|} {D_f < -\beta} 
\end{equation}
where $r$ is the final reward awarded to the actor and $\beta$ is a user defined constant between $0$ to $100$. Similarly the action sequence in every time step is determined by,
\begin{equation}
    \label{eq:scaf_action}
    a = \twopartdef { a_r } {D_f \geq -\beta} {a_p} {D_f < -\beta} 
\end{equation}
Thus, the experience tuple $<s,a,r,s'>$ is formed using the above formulation. This fastens the process of learning, as the probability of initial randomization of the actions space is minimized. The directions desired by the user (as PMF) gets embedded into the actor as the training starts. Thus, allowing it to reflect upon its action by using PFM as a scaffold.

\section{Numerical simulations}
\label{sec:eval}

In this section, the working of the SR2L framework is presented. A trained evader is used to demonstrate the escape against multiple defenders. Following that, convergence studies of both SR2L and independent actor-critic  (IAC) are presented. It is then used to evaluate SR2L training capabilities by comparing it with IAC as the baseline. Finally, a performance evaluation is done by comparing the performance metrics of SR2L with that of IAC and PFM.

\subsection{SR2L for confinement escape problem}
\label{sec:sr2l_performance}

\begin{figure}
\centering
\includegraphics[clip,trim=0mm 0mm 0mm 3mm,width=1\columnwidth]{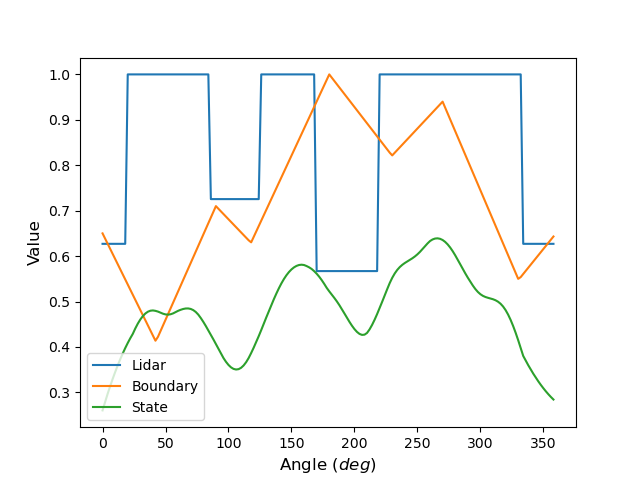}
\caption{The state and individual data representation of Lidar and boundary information at t = $t_{max}/2$.}.
\label{state}
\end{figure}

Consider a confinement region $A$, having rectangular boundary of dimension $200$ m $\times$ $200$ m. Let the number of pursuers patrolling the area be $30$ (i.e, $n = 30$). The evader is initialized randomly within an area $\Omega$ of dimension $20$ m $\times$ $20$ m  near the origin. The pursuers are initialized randomly within the confinement boundary and outside of $\Omega$. The evader's initial velocity is taken to be zero, with the maximum velocity limit being $V_e^{max} = 15$ m/s. The pursuers are initialized with a random velocity between $V_p^{min} = 5$ m/s and $V_p^{max} = 10$ m/s. The sensor radius of evader and pursuer is $15$ m and $10$ m, respectively.
The sequence of actions by the evader in different time steps is given in Fig. \ref{fig:action_sequence}. 
Here, the green plot represents the evader's path, and the blue arcs represent the Lidar point cloud data.
The point cloud data representation, boundary information representation and the state of the evader at time step = 25 (Fig. \ref{fig:a}) is given in Fig. \ref{state}. The axis corresponds to the direction angle, considering the evaders heading as $0$ deg. 
Here, the action space of dimension $2$ is considered. This action space consists of the evader's velocity along the $x$ and $y$ axis for motion control.
The evader's trajectory in Fig. \ref{fig:action_sequence} shows the evader escaping from the nearest boundary point. In Fig. \ref{fig:c}, the evader is shown to have crossed the line of $x = -100$, thus successfully escaping the confinement. Even though multiple pursuers detect the evader and start chasing, it was able to change its course multiple times, as shown in Fig. \ref{fig:c}, for avoiding capture while achieving a successful escape. It is also aided by the fact that the max speed ratio of $V_e^{max}/V_p^{max} > 1$ is considered. For visualization, the video of CEP using SR2L is provided in the following link: {\bf https://bit.ly/3nSrI1I}

\begin{figure}
\centering

\subcaptionbox{independent actor-critic  training framework.\label{fig:comp_a}}
{\includegraphics[clip,trim=0mm 0mm 0mm 14mm,width=1\columnwidth]{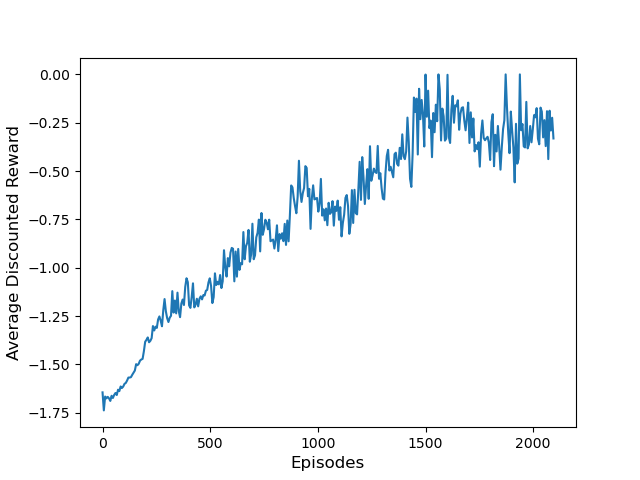}}

\subcaptionbox{SR2L training framework.\label{fig:comp_b}}
{\includegraphics[clip,trim=0mm 0mm 0mm 10mm,width=1\columnwidth]{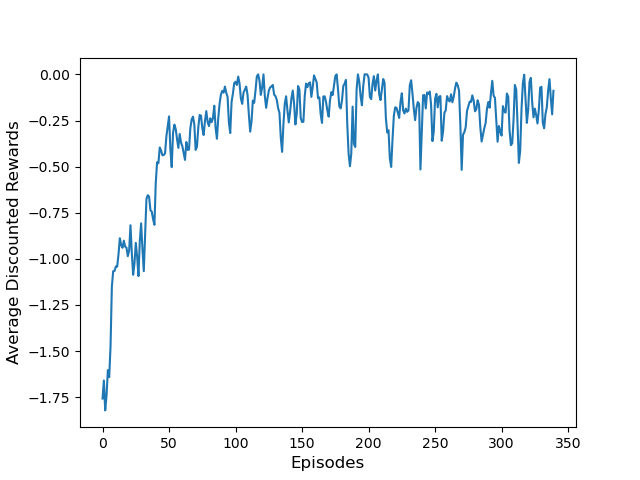}}


\caption{The convergence plots for training actor-critic networks for confinement escape problem.}
\label{convergence_sr2l}
\end{figure}


\renewcommand{\arraystretch}{1.5}
\begin{center}

\begin{table}[]
\setlength\tabcolsep{3pt}
\scriptsize
\centering
\begin{tabular}[c]{ | l | c | c | c | c | c | c | c | c | c | c |} 
\hline
No. of pursuers & \multicolumn{2}{c|}{\textbf{10}} & \multicolumn{2}{c|}{\textbf{20}} & \multicolumn{2}{c|}{\textbf{30}} & \multicolumn{2}{c|}{\textbf{40}} & \multicolumn{2}{c|}{\textbf{50}} \\
\hline
 & \textbf{\%} & \textbf{time} & \textbf{\%} & \textbf{time} & \textbf{\%} & \textbf{time} & \textbf{\%} & \textbf{time} & \textbf{\%} & \textbf{time}\\
\hline
V' = 1, R' = 1 & 100 & 76.3 & 97 & 96.5 & 91 & 132.8 & 86 & 162.4 & 77 & 199.2\\
\hline
V' = 1, R' = 2 & 100 & 78.3 & 100 & 88.9 & 96 & 121.8 & 91 & 146.4 & 82 & 185.6\\
\hline
V' = 1.5,R' = 1.5 & 100 & 66.8 & 100 & 79.1 & 100 & 100.4 & 93 & 132.6 & 85 & 151.1\\
\hline
V' = 2, R' = 1 & 100 & 59.7 & 100 & 70.5 & 100 & 89.6 & 96 & 112.8 & 91 & 136.8\\
\hline
V' = 2, R' = 2 & 100 & 64.2 & 100 & 76.9 & 100 & 91.5 & 100 & 108.6 & 98 & 128.9\\
\hline
\end{tabular}
\caption{Performance of trained SR2L policy in different environment parameters.}
\label{tab:table1}
\end{table}
\end{center}

\subsection{Convergence Analysis of SR2L and  IAC}

The environment used in section \ref{sec:sr2l_performance} is adopted for the convergence study. During training, for a given state $s$, the evader must learn to choose an action $a$ to maximize the reward $r$. The evader is trained using two different methods, a) independent actor-critic framework, b) SR2L framework.
In the former, the Actor-Critic is trained by using the reward generated by Eqn \ref{eq:reward}. Here, the time step $dt = 0.1$ units and termination time $t^{max} = 300$ units is taken. The constants $K_s = 1$, $R_e = 40$, $d_b^{max} = 200$ and $\gamma = 0.001$ are considered. The weights $W_L = W_B = 1$ for the state generation are used. The convergence plot of this learning process is given in Fig. \ref{fig:comp_a}. For the second case, same constant values are used. However, the action is chosen by using Eqn. \eqref{eq:scaf_action}, and the corresponding reward is awarded using Eqns. \eqref{eq:reward} \& \eqref{eq:scaf_reward}. The convergence plot for this case is given in Fig. \ref{fig:comp_b}.

The figures show that the learning process using the independent actor-critic framework converges after approximately 1500 episodes. Whereas in the case of SR2L, it converges after approximately 100 episodes. Hence, it achieves a convergence that is approximately $15$ times faster than IAC. Also, SR2L converges to a higher approximate average reward of $-0.15$ as compared to $-0.25$ for IAC. 


\subsection{Monte-Carlo Performance Evaluation of SR2L}


\begin{figure}
\centering

\subcaptionbox{Average time taken to escape confinement.\label{fig:sr2l_vs_pfm_a}}
{\includegraphics[clip,trim=0mm 0mm 0mm 14mm,width=1\columnwidth]{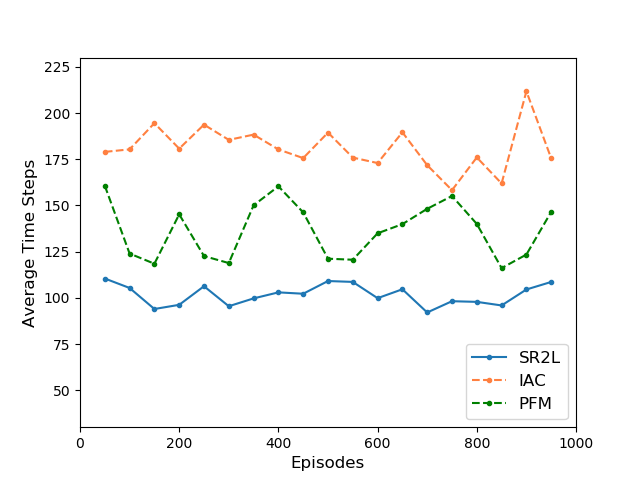}}

\subcaptionbox{Average reward while attempting escape.\label{fig:sr2l_vs_pfm_b}}
{\includegraphics[clip,trim=0mm 0mm 0mm 10mm,width=1\columnwidth]{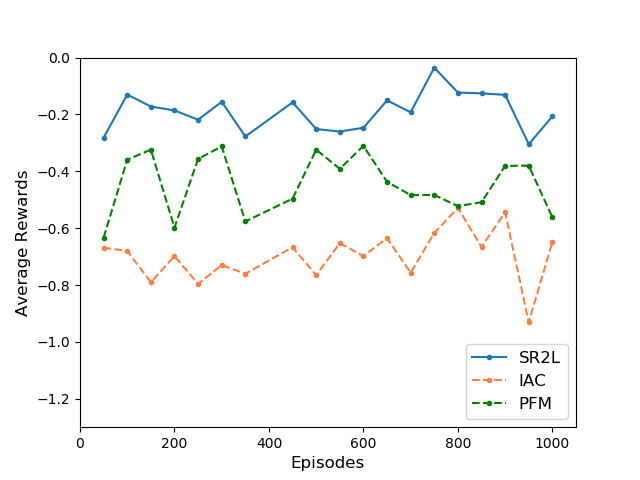}}

\caption{The comparison plots of SR2L against PFM.}
\label{time_sr2l}
\end{figure}

For the Monte-Carlo simulation study, the same environment presented in section \ref{sec:sr2l_performance} is used. The simulation was conducted for three different cases of evader, i.e., a) trained using SR2L for 100 episodes, b) trained using IAC for 1000 episodes and c) using the potential field method as motion planner. The Monte-Carlo simulation study was carried out by playing the game for $1000$ episodes with random initialization of evader and pursuers. The average time taken by the three types of evaders to escape confinement per $100$ game is plotted in fig. \ref{fig:sr2l_vs_pfm_a}. It shows that the average time is taken for escape using SR2L always lower than that of PFM. For SR2L, $100$ times steps approximately are required while IAC and PFM require $185$ and $140$ time steps, respectively.

Also, the performance comparison between an SR2L trained evader for $100$ episodes and an IAC trained evader for $1000$ episodes is done. In the fig. \ref{fig:sr2l_vs_pfm_a} it can be seen that average time required for completion for IAC is $180$ as compared to $100$ for SR2L. The state information in SR2L uses time factor and distance to boundary, which helps evader using SR2L policy to escape faster than PFM.  Also, the faster convergence during training provides SR2L with an edge over IAC. Please note that if given enough time for training, the IAC will also reach optimal reward conditions as the primary reward function used in both cases are the same. 

During these simulations, the evader's reward was also recorded for all $1000$ episodes. The average reward achieved by both evaders per $100$ episodes is plotted in Fig. \ref{fig:sr2l_vs_pfm_b}. As the PFM is used as a scaffold in the SR2L framework, so the latter always has a higher average reward for CEP. 
The average reward achieved by SR2L is consistently higher as compared to IAC and PFM. SR2L achieves an approximate reward of $-0.2$ per 100 episodes. Whereas in the case of IAC and PFM, it is $-0.75$ and $-0.5$, respectively. These results also reflect a learning algorithm's ability to capture more features from the states than a handcrafted PFM algorithm.

For further evaluation, Monte-Carlo simulation was carried out with varying total number of pursuers, the maximum speed ratio $V' = V_e^{max}/V_p^{max}$ and the ratio of sensing radius $R' = R_e/R_p$. The results have been tabulated and presented in table \ref{tab:table1}. It shows the percentage of successful escape in corresponding scenarios along with the average time of escape. Consider the worst-case scenario, where $V' = 1$ and $R' = 1$ is used, and $50$ robots are patrolling the area. With $10m$ sensor radius, considering $20\%$ of overlap of sensing area among the pursuers, at any given time, $40\%$ of the area is under the coverage of the pursuers. Even in such a scenario, 77\% of the time evader escapes successfully. Here, the failure cases include 16 cases of evader being captured and 7 cases exceeding the time limit for escape. The results show that the evader trained using SR2L can escape with varying environmental parameters with different numbers of robots, thus highlighting its scalability.

\section{Conclusions}
\label{sec:conclusion}

This paper introduces the scaffolding reflection in reinforcement learning framework to solve the confinement escape problem for an evader. 
It uses a motion planner to train an evader to escape the confinement against multiple dynamic pursuers while using lesser resources and time. The framework is independent of the efficiency of the motion planner used. The reward function consisting of information about the vicinity, boundary, and time elapsed is designed to emulate the true objective. The potential field method is used as a scaffold for the actor-critic network to observe, compare and learn. As SR2L considers an observable environment in the vicinity, it is computationally efficient. However, the issue of dimensionality can still arise because of the number of pursuers. Hence, a state representation is designed for the learning algorithm to capture the environment's spatial features to avoid dimensionality issues. The convergence studies show that the SR2L converges $15$ times faster than independent actor-critic  method. Also, during training, the SR2L method converges to a higher average reward that is approximately 1.6 times that of the IAC method. Monte-Carlo simulations show that the approximate average reward achieved by SR2L is 3.75 times higher than the IAC and 2.5 times higher than PFM. Also, the total time required for IAC and PFM to achieve the same objective value is approximately 1.4 times more than SR2L.



\bibliographystyle{IEEEtran}
\bibliography{main}

\end{document}